%% file: main.tex
\documentclass[11pt,a4paper]{article}
\usepackage[acceptedWithA, hyperref]{tacl2018v2}
\usepackage{times}
\usepackage{latexsym}
\usepackage[T1]{fontenc}

\usepackage{amsmath,amsfonts,amssymb}
\usepackage{amsthm}
\usepackage{proof}

\usepackage{xspace}
\usepackage{comment}
\usepackage{framed}
\usepackage{tikz}
\usetikzlibrary{patterns}
\usepackage{tikzscale}
\usepackage{pgfplots}
\usepackage{pgfplotstable}
\pgfplotsset{compat=1.14}
\usepgfplotslibrary{groupplots}
\usepackage{booktabs}
\usepackage{verbatimbox}
\usepackage{fixltx2e}
\usepackage{subcaption}
\usepackage{float}
\usepackage{caption}
\usepackage[inline]{enumitem}
\usepackage{xspace}
\usepackage{graphicx}
\usepackage{tabularx}
\usepackage{scalefnt}
\usepackage{multirow}
\usepackage{graphicx}
\usepackage{tabularx}

\usepgfplotslibrary{groupplots}
\usetikzlibrary{matrix}

\newcommand\smaller[2][0.85]{{\scalefont{#1}#2}}
\newtheorem{theorem}{Theorem}
\newtheorem{lemma}{Lemma}

\newtheorem{corollary}{Corollary}
\theoremstyle{definition}
\newtheorem{definition}{Definition}[section]
\theoremstyle{remark}
\newtheorem*{remark}{Remark}

\newcommand{\set}[1]{\ensuremath \mathcal{#1}}
\newcommand{\clstok}{\textsc{[cls]}\xspace}
\newcommand{\bert}{\smaller{BERT}\xspace}
\newcommand{\msmarco}{{\small{MS MARCO}}\xspace}
\newcommand{\treccar}{{\small{TREC-CAR}}\xspace}

\newcommand{\printfnsymbol}[1]{%
  \textsuperscript{\@fnsymbol{#1}}%
}
\newcommand{\commentout}[1]{}

\setlist[itemize]{parsep=1pt,itemsep=1pt,topsep=5pt,leftmargin=\parindent}

\newif\iftaclinstructions
\taclinstructionsfalse 
\iftaclinstructions
\renewcommand{\confidential}{}
\renewcommand{\anonsubtext}{(No author info supplied here, for consistency with
TACL-submission anonymization requirements)}
\newcommand{\instr}
\fi

\taclpubformattrue
\iftaclpubformat 

\else

\fi

\title{Sparse, Dense, and Attentional Representations for Text Retrieval}

\author{Yi Luan\thanks{{ } equal contribution}, Jacob Eisenstein$^*$,  Kristina Toutanova$^*$,  Michael Collins \\
Google Research \\
\texttt{\large \{luanyi, jeisenstein, kristout, mjcollins\}@google.com} \\
}

\date{}

\newcommand{\reals}[0]{\ensuremath \mathbb{R}}
\newcommand{\vocsize}[0]{\ensuremath v}

\newcommand{\maxtoks}[0]{\ensuremath T}
\newcommand{\embsize}[0]{\ensuremath k}

\newcommand{\iprod}[2]{\ensuremath \langle #1, #2 \rangle}
\newcommand{\term}[1]{\textbf{#1}}

\newcommand{\toksfont}[1]{\ensuremath \textrm{#1}}
\newcommand{\xtoks}[0]{\ensuremath \toksfont{x}}
\newcommand{\ytoks}[0]{\ensuremath \toksfont{y}}
\newcommand{\tfidf}{\smaller{TF-IDF}\xspace}

\newcommand{\newtext}[1]{\textcolor{black}{#1}}
\newcommand{\newtextfinal}[1]{\textcolor{black}{#1}}
\newcommand{\newtextbresub}[1]{\textcolor{black}{#1}}

\input{commands}

\begin{document}
\maketitle
\begin{abstract}
\input{abstract}
\end{abstract}

\input{introduction}
\input{de_theory_from_error_bound}
\input{multivector}
\input{experiments}

\input{experiments-retrieval.tex}

\input{related}
\input{conclusion}
\input{acknowledgments} 

\bibliography{cites}
\bibliographystyle{acl_natbib}

\appendix
\input{proofs}
\input{tables/ICT_by_length}

\end{document}

%% file: commands.tex
\definecolor{red1}{rgb}{10,0,1}

\newcommand{\addrecallbert}[0]{
    \addplot[mark=diamond*,red, semithick, mark options={mark size=2}]
        table [x=index, y=bert-avg-pool-768, col sep=comma] 
}

\newcommand{\addrecallplotdethreetwo}[0]{
    \addplot[mark=o,blue!30!cyan, semithick, mark options={mark size=2}]
        table [x=index, y=deproj32, col sep=comma] 
}

\newcommand{\addrecallplotdesixfour}[0]{
    \addplot[mark=x,blue!50!cyan, semithick, mark options={mark size=2}]
        table [x=index, y=deproj64, col sep=comma] 
}
\newcommand{\addrecallplotdeonetwoeight}[0]{
    \addplot[mark=square,blue!70!cyan, semithick, mark options={mark size=2}]
        table [x=index, y=deproj128, col sep=comma] 
}

\newcommand{\addrecallplotdefivetwelve}[0]{
    \addplot[mark=triangle,blue!70!cyan, semithick, mark options={mark size=2}]
        table [x=index, y=deproj512, col sep=comma] 
}

\newcommand{\addrecallplotdesevensixeight}[0]{
    \addplot[mark=diamond,blue!70!black, semithick, mark options={mark size=2}]
        table [x=index, y=deproj768, col sep=comma] 
}

\newcommand{\addrecallplotqonedeight}[0]{
    \addplot[mark=square,red, semithick, mark options={mark size=2}]
        table [x=index, y=q1d8proj768, col sep=comma] 
}

\newcommand{\addrecallplotqonedeightsixfour}[0]{
    \addplot[mark=o,red!70!yellow, semithick, mark options={mark size=2}]
        table [x=index, y=q1d8proj64, col sep=comma] 
}

\newcommand{\addrecallplotbmone}[0]{
    \addplot[mark=diamond*,gray!70!black, semithick, mark options={mark size=2}]
        table [x=index, y=bm25uni, col sep=comma] 
}
\newcommand{\addrecallplotbmtwo}[0]{
    \addplot[mark=*,gray!70!black, semithick, mark options={mark size=2}]
        table [x=index, y=bm25bi, col sep=comma] 
}

\newcommand{\addrecallplotbmqonedeightfull}[0]{
    \addplot[mark=star,green!60!black, semithick, mark options={mark size=2}]
        table [x=index, y=q1d8proj768_hybrid, col sep=comma] 
}

\newcommand{\addrecallplotbmqonedeightfullbi}[0]{
    \addplot[mark=star,red!60!blue, semithick, mark options={mark size=2}]
        table [x=index, y=q1d8proj768_hybrid_bi, col sep=comma] 
}

\newcommand{\addmsmarcobm}[0]{
    \addplot[mark=square,gray!70!black, semithick, mark options={mark size=2}]
        table [x=index, y=BM25, col sep=comma] 
}

\newcommand{\addmsmarcodebert}[0]{
    \addplot[mark=star,blue, semithick, mark options={mark size=2}]
        table [x=index, y=DE-BERT, col sep=comma] 
}

\newcommand{\addmsmarcomebert}[0]{
    \addplot[mark=diamond,orange!30!red, semithick, mark options={mark size=2}]
        table [x=index, y=ME-BERT, col sep=comma] 
}

\newcommand{\addmsmarcohybrid}[0]{
    \addplot[mark=o,green!60!black, semithick, mark options={mark size=2}]
        table [x=index, y=Hybrid, col sep=comma] 
}

\newcommand{\addmsmarcodeepct}[0]{
    \addplot[mark=triangle,purple, semithick, mark options={mark size=2}]
        table [x=index, y=DeepCT, col sep=comma] 
}

\newcommand{\addrecallplotca}[0]{
    \addplot+[mark size=2pt, mark=otimes*, color=purple, dashed, semithick ]
        plot [error bars/.cd, y dir = both, y explicit]
        table [x=index, y=fullcross, col sep=comma] 
}

\newcommand{\crossatt}{\smaller{Cross-Attention\xspace}}

\newcommand{\bmtwo}{\smaller{BM25}-bi\xspace}

\newcommand{\bm}{\smaller{BM25}\xspace}

\newcommand{\bmone}{\smaller{BM25}-uni\xspace}
\newcommand{\hybrid}{\smaller{HYBRID}\xspace}

\newcommand{\de}{\smaller{DE-BERT}\xspace}

\newcommand{\qonedeight}{\smaller{ME-BERT}\xspace}



%% file: abstract.tex
Dual encoders perform retrieval by encoding documents and queries into dense low-dimensional vectors, scoring each document by its inner product with the query. 
We investigate the capacity of this architecture relative to sparse bag-of-words  models and attentional neural networks.
\newtext{Using both theoretical and empirical analysis, we establish connections between the encoding dimension, the margin between gold and lower-ranked documents, and the document length, suggesting limitations 
in the capacity of fixed-length encodings to support \newtextbresub{precise} retrieval of long documents.}
Building on these insights, we propose a simple neural model that combines the efficiency of dual encoders with some of the expressiveness of more costly attentional architectures, and explore sparse-dense hybrids to capitalize on the precision of  sparse retrieval.
These models outperform strong alternatives in large-scale retrieval.

%% file: introduction.tex
\section{Introduction}
Retrieving relevant documents is a core task for language technology, and is a component of applications such as information extraction and question answering ~\cite[e.g.,][]{narasimhan2016improving,kwok2001scaling,voorhees_2001}.
While classical information retrieval has focused on heuristic weights for sparse bag-of-words representations~\cite{sparck1972statistical}, more recent work has adopted a two-stage retrieval and ranking pipeline, where a large number of documents are retrieved using sparse high dimensional query/document representations, and are further reranked with learned neural models~\cite{mitra2018an}.
This two-stage approach has achieved state-of-the-art results on IR benchmarks~\cite{passage-rerank, bert-adhoc, nogueira2019multistage}, especially since sizable annotated data has become available for training deep neural models~\cite{treccar18,craswell2020overview}. However, this pipeline suffers from a strict upper bound imposed by any recall errors in the first-stage retrieval model: for example, the recall@1000 for \bm reported by \citet{alibaba20} is 69.4.

\input{figures/bertvsbm25new.tex}

A promising alternative is to perform first-stage retrieval using learned dense low-dimensional encodings of documents and queries~\cite{huang2013learning,reimers-gurevych-2019-sentence,gillick-etal-2019-learning,karpukhin2020dense}. The dual encoder model scores each document by the inner product between its encoding and that of the query. Unlike full attentional architectures, which require extensive computation on each candidate document, the dual encoder can be easily applied to very large document collections thanks to efficient algorithms for inner product search; unlike untrained sparse retrieval models, it can exploit machine learning to generalize across related terms.

\newtextbresub{To assess the relevance of a document to an information-seeking query, models must \emph{both} (\text{i}) check for precise term overlap (for example, presence of key entities in the query) and (\textit{ii}) compute semantic similarity generalizing across related concepts.  Sparse retrieval models excel at the first sub-problem, while learned dual encoders can be better at the second.}
Recent history in NLP might suggest that learned dense representations should always outperform sparse features overall, but this is not necessarily true: as shown in \autoref{fig:1-retrieve-passage-contain-q}, the \bm  model~\cite{robertson2009probabilistic} can outperform a dual encoder based on  \bert, particularly on longer documents and on a task that requires precise detection of word overlap.\footnote{See \autoref{sec:experiments} for experimental details.} This raises questions about the  limitations of dual encoders, and the circumstances in which these powerful models do not yet reach the state of the art.
Here we explore these questions using both theoretical and empirical tools, and propose a new architecture that leverages the strengths of dual encoders while avoiding some of their weaknesses.

We begin with a theoretical investigation of compressive dual encoders --- dense encodings whose dimension is below the vocabulary size --- and analyze their ability to preserve distinctions made by sparse bag-of-words retrieval models, which we term their \term{fidelity}. \newtextbresub{Fidelity is important for the sub-problem of detecting precise term overlap, and is a tractable proxy for capacity.}
Using the theory of dimensionality reduction, we relate fidelity to the normalized margin between the gold retrieval result and its competitors, and show that this margin is in turn related to the length of documents in the collection.  
We validate the theory with an empirical investigation of the effects of \newtextfinal{random projection} compression on sparse \bm retrieval using queries and documents from \treccar, a recent IR benchmark ~\cite{treccar18}.

Next, we offer a multi-vector encoding model, which is \newtextfinal{computationally feasible for retrieval} like the dual-encoder architecture and achieves significantly better quality. A simple hybrid that interpolates models based on dense and   sparse representations leads to further improvements.

We compare the performance of dual encoders, multi-vector encoders, and their sparse-dense hybrids with classical sparse retrieval models and attentional neural networks, as well as state-of-the-art published results where available. Our evaluations include open retrieval benchmarks (\msmarco passage and document), and passage retrieval for question answering (Natural Questions).
We confirm prior findings that full attentional architectures excel at reranking tasks, but are not efficient enough for large-scale retrieval. Of the more efficient alternatives, the hybridized multi-vector encoder is at or near the top in every evaluation, outperforming state-of-the-art retrieval results in \msmarco. Our code is publicly available at \url{https://github.com/google-research/language/tree/master/language/multivec}.

%% file: figures/bertvsbm25new.tex
\begin{figure}

\begin{tikzpicture}
    \pgfplotsset{footnotesize,samples=10}
    \begin{groupplot}[group style = {group size = 1 by 1, horizontal sep = 50pt}, width = 5.5cm, height = 3.75cm]
    \nextgroupplot[
            xticklabels from table={passage_sizes.dat}{Label},
            xticklabel style={align=center,font=\small},
            legend style = { column sep = 10pt, legend columns = -1, legend to name = grouplegend,},
            legend columns=2,
            every tick label/.append style={font=\small},
            legend style={column sep=0.2cm},
            legend style={anchor=south},
            legend style={at={(0.5, 0.0), anchor=south}, font=\large},
            ymin=0,ymax=100,
            enlarge y limits=true,
xlabel={passage length}, ylabel={recall@1},
            ]
    \addrecallplotbmone{figures/data/results/unmasked_retrieve_efficient_test.csv};\label{addrecallplotbmone}    
        \addrecallplotbmtwo{figures/data/results/unmasked_retrieve_efficient_test.csv};\label{addrecallplotbmtwo} 
      \addrecallplotdesevensixeight{figures/data/results/unmasked_retrieve_efficient_test.csv};\label{addrecallplotdesevensixeight}
        \addrecallbert{figures/data/results/unmasked_retrieve_efficient_test.csv};\label{addrecallbert}
\coordinate (top) at (rel axis cs:0,1);
 
\coordinate (bot) at (rel axis cs:1,0);
    \end{groupplot}
    
    \path (top)--(bot) coordinate[midway] (group center);
        
  \matrix[
      matrix of nodes,
      anchor=west,
      inner sep=0.2em,
    ]at([right=0.2em,inner sep=0pt]group center -| current bounding box.east)
    {\ref{addrecallplotbmone}& \small{\bmone} \\
    \ref{addrecallplotbmtwo}& \small{\bmtwo} \\
        \ref{addrecallplotdesevensixeight}& \small{\de}-\small{768}  \\
        \ref{addrecallbert}& \small{\smaller{BERT-init}} \\
            };

\end{tikzpicture}

    \caption{\small Recall@1 for retrieving passage containing a query from three million candidates. The figure compares a fine-tuned {\bert}-based dual encoder ({\de-{\smaller 768}}), an off-the-shelf {\bert}-based encoder with average pooling (\smaller{BERT-init}), and sparse term-based retrieval (\bm), while binning passages by length.
    }
    \label{fig:1-retrieve-passage-contain-q}
\end{figure}
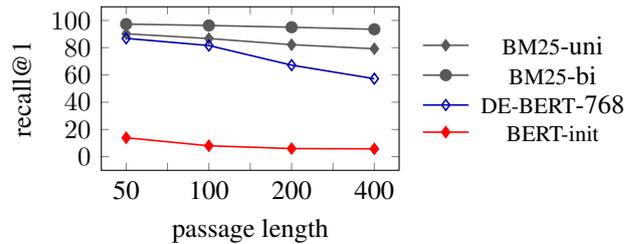

%% file: de_theory_from_error_bound.tex
\section{Analyzing dual encoder fidelity}
\label{sec:theory}
A query or a document is a sequence of words drawn from some vocabulary $\set{V}$. 
Throughout this section we assume a representation of queries and documents typically used in sparse bag-of-words models: each query $q$ and document $d$ is a vector in $\reals^\vocsize$ where $\vocsize$ is the vocabulary size. We take the inner product $\iprod{q}{d}$ to be the relevance score of document $d$ for query $q$. This framework accounts for a several well-known ranking models, including boolean inner product, \tfidf, and \bm. 

We will compare sparse retrieval models with compressive dual encoders, for which we write $f(d)$ and $f(q)$ to indicate compression of $d$ and $q$ to $\reals^{\embsize}$, with $\embsize \ll \vocsize,$ and where $\embsize$ does not vary with the document length. For these models, the relevance score is the inner product $\iprod{f(q)}{f(d)}$. (In \autoref{sec:multivector}, we consider encoders that apply to sequences of tokens rather than vectors of counts.)

A fundamental question is how the capacity of dual encoders varies with the embedding size $\embsize$. In this section we focus on the related, more tractable notion of fidelity: how much can we compress the input while maintaining the ability to mimic the performance of bag-of-words retrieval? We explore this question mainly through the encoding model of random projections, but also discuss more general dimensionality reduction in \autoref{sec:general-limits}.

\subsection{Random projections}
To establish baselines on the fidelity of compressive dual encoder retrieval, we now consider encoders based on \term{random projections}~\cite{vempala2004random}.
The encoder is defined as $f(x) = Ax,$ where $A \in \reals^{k \times v}$ is a random matrix.
In \term{Rademacher embeddings}, each element $a_{i,j}$ of the  matrix $A$ is sampled with equal probability from two possible values: $\{-\frac{1}{\sqrt{\embsize}},\frac{1}{\sqrt{\embsize}}\}$.
In \term{Gaussian embeddings}, each $a_{i,j} \sim N(0,\embsize^{-1/2})$. 
A pairwise ranking error occurs when $\iprod{q}{d_1} > \iprod{q}{d_2}$ but $\iprod{Aq}{Ad_1} < \iprod{Aq}{Ad_2}$.
Using such random projections, it is possible to bound the probability of any such pairwise error in terms of the embedding size.

\newcommand{\nmarg}[0]{\ensuremath \mu}
\begin{definition}
For a query $q$ and pair of documents $(d_1, d_2)$ such that $\iprod{q}{d_1} \geq \iprod{q}{d_2}$, the \term{normalized margin} is defined as,
$\nmarg(q, d_1, d_2) = \frac{\iprod{q}{d_1 - d_2}}{||q|| \times ||d_1 - d_2||}$.
\end{definition}

\begin{lemma}
\label{lem:ranking-error-rp}
Define a matrix $A \in \reals^{k \times d}$ of Gaussian or Rademacher embeddings.  
Define vectors $q, d_1, d_2$ such that $\nmarg(q, d_1, d_2) = \epsilon > 0$. 
A ranking error occurs when $\iprod{Aq}{Ad_2} \geq \iprod{Aq}{Ad_1}$.
If $\beta$ is the probability of such an error then, 
\begin{equation}
\beta \leq 4 \exp\left(-\frac{\embsize}{2}(\epsilon^2/2 - \epsilon^3/3)\right).
\label{eq:bd-bound-in-lemma}
\end{equation}
\end{lemma}

The proof, which builds on well-known results about random projections, is found in \autoref{sec:proof_RP_error_bound}. By solving (\ref{eq:bd-bound-in-lemma}) for $\embsize$, we can derive an embedding size that guarantees a desired upper bound on the pairwise error probability, 
\begin{equation}
    \embsize \geq 2 (\epsilon^2/2 - \epsilon^3/3)^{-1} \ln \frac{4}{\beta}.
\end{equation}

It is convenient to derive a simpler but looser quadratic bound (proved in \autoref{sec:proof-quad-bound}):
\begin{corollary}
\label{cor:quad-bound}
Define vectors $q, d_1, d_2$ such that $\epsilon = \nmarg(q, d_1,d_2) > 0$. If $A \in \reals^{\embsize \times \vocsize}$ is a matrix of random Gaussian or Rademacher embeddings such that $\embsize > 12 \epsilon^{-2} \ln \frac{4}{\beta}$, then $\Pr(\iprod{Aq}{Ad_1} \leq \iprod{Aq}{Ad_2}) \leq \beta$.
\end{corollary}

\paragraph{On the tightness of the bound.} Let $k^*(q,d_1,d_2)$ denote the lowest dimension Gaussian or Rademacher random projection following the definition in Lemma 1, for which $\Pr(\iprod{Aq}{Ad_1} < \iprod{Aq}{Ad_2}) \leq \beta$, for a given document pair $(d_1,d_2)$ and query $q$ with normalized margin $\epsilon$. Our lemma places an upper bound on $k^*$, saying that $k^*(q,d_1,d_2) \leq 2 (\epsilon^2/2 - \epsilon^3/3)^{-1} \ln \frac{4}{\beta}$. Any $k \geq k^*(q, d_1, d_2)$ has sufficiently low probability of error, but lower values of $k$ could potentially also have the desired property. Later in this section we perform empirical evaluation to study the tightness of the bound; although theoretical tightness (up to a constant factor) is suggested by results on the optimality of the distributional Johnson-Lindenstrauss lemma \cite{jl84,jw13,kmn11}, here we study the question only empirically.

\newcommand{\recallthresh}[0]{\ensuremath r}
\subsubsection{Recall-at-$\recallthresh$}
In retrieval applications, it is important to return the desired result within the top $\recallthresh$ search results. For query $q$, define $d_1$ as the document that maximizes some inner product ranking metric. The probability of returning $d_1$ in the top $\recallthresh$ results after random projection can be bounded by a function of the embedding size and normalized margin:

\begin{lemma}
\label{lem:recall-at-r}
Consider a query $q$, with target document $d_1$, and document collection ${\cal D}$ that excludes $d_1$, and such that $\forall d_2 \in {\cal D}, \nmarg(q, d_1, d_2) > 0$. Define $\recallthresh_0$ to be any integer such that $1 \leq \recallthresh_0 \leq |{\cal D}|$. Define $\epsilon$ to be the $\recallthresh_0$'th smallest normalized margin $\nmarg(q, d_1, d_2)$ for any $d_2 \in {\cal D}$, and for simplicity assume that only a single document $d_2 \in {\cal D}$ has $\nmarg(q, d_1, d_2) = \epsilon$.\footnote{The case where multiple documents are tied with normalized margin $\epsilon$ is straightforward but slightly complicates the analysis.}

Define a matrix $A \in \reals^{\embsize \times d}$ of Gaussian or Rademacher embeddings. Define $R$ to be a random variable such that $R = |\{d_2 \in {\cal D}: \langle A q, A d_1 \rangle  \leq \langle Aq,  A d_2 \rangle\}|$, and let $C = 4 (|{\cal D}| - \recallthresh_0 + 1)$. Then 
\begin{equation*}
\Pr(R \geq \recallthresh_0) \leq C \exp \left( - \frac{\embsize}{2} (\epsilon^2/2 - \epsilon^3/3)\right).
\end{equation*}
\end{lemma}

The proof is in \autoref{sec:proof-recall-at-r}.
A direct consequence of the lemma is that to achieve recall-at-$\recallthresh_0=1$ for a given $(q, d_1, {\cal D})$ triple with probability $\geq 1 - \beta$, it is sufficient to set 
\begin{equation}
\embsize \geq \frac{2}{\epsilon^2/2 - \epsilon^3/3} \ln \frac{4 (|{\cal D}| - \recallthresh_0 + 1)}{\beta},
\end{equation}
where $\epsilon$ is the $\recallthresh_0$'th smallest normalized margin. 

As with the bound on pairwise relevance errors in Lemma 1, Lemma 2 implies an upper bound on the minimum random projection dimension $k^*(q,d_1,{\cal D})$ that recalls $d_1$ in the top $r_0$ results with probability $\geq 1 -\beta$.  Due to the application of the union bound and worst-case assumptions about the normalized margins of documents in ${\cal{D}}_\epsilon$, this bound is potentially loose. Later in this section we examine the empirical relationship between maximum document length, the distribution of normalized margins, and  $k^{*}$.

\subsubsection{Application to Boolean inner product}
Boolean inner product is a retrieval function in which $d, q \in \{0, 1\}^{\vocsize}$ over a vocabulary of size $\vocsize$, with $d_i$ indicating the presence of term $i$ in the document (and analogously for $q_i$). The relevance score $\iprod{q}{d}$ is then the number of terms that appear in both $q$ and $d$. For this simple retrieval function, it is possible to compute an embedding size that guarantees a desired pairwise error probability over an entire dataset of documents. 
\begin{corollary}
For a set of documents $\set{D} = \{d \in \{0, 1\}^{\vocsize}\}$ and a query $q \in \{0, 1\}^{\vocsize}$,
let $L_D = \max_{d \in \set{D}} ||d||^2$ and $L_Q = ||q||^2$. Let 
$A \in \reals^{\embsize \times \vocsize}$ be a matrix of random Rademacher or Gaussian embeddings such that $\embsize \geq 24 L_Q L_D \ln \frac{4}{\beta}.$ Then for any $d_1, d_2 \in \set{D}$ such that $\iprod{q}{d_1} > \iprod{q}{d_2}$, the probability that $\iprod{Aq}{Ad_1} \leq \iprod{Aq}{Ad_2}$ is $\leq \beta$.
\label{cor:boolean-inner-product}
\end{corollary}

The proof is in \autoref{sec:proof-boolean-ip}.
The corollary shows that for boolean inner product ranking, we can guarantee any desired error bound $\beta$ by choosing an embedding size $\embsize$ that grows linearly in $L_D$, the number of unique terms in the longest document. 

\subsubsection{Application to \tfidf and \bm}
\newtext{Both \tfidf~\cite{sparck1972statistical} and \bm~\cite{robertson2009probabilistic} can be written as inner products between bag-of-words representations of the document and query as described earlier in this section.
Set the query representation $\tilde{q}_i = q_i \times \textsc{idf}_i$, where $q_i$ indicates the presence of the term in the query and $\textsc{idf}_i$ indicates the inverse document frequency of term $i$. The \tfidf score is then $\iprod{\tilde{q}}{d}$. For \bm, we define $\tilde{d} \in \reals^{\vocsize}$, with each $\tilde{d}_i$ a function of the count $d_i$ and the document length (and hyperparameters); $\text{\bm}(q,d)$ is then ${\iprod{\tilde{q}}{\tilde{d}}}$.} Due to its practical utility in retrieval, we now focus on \bm.

\begin{figure}
    \centering
    \includegraphics[width=.9\columnwidth]{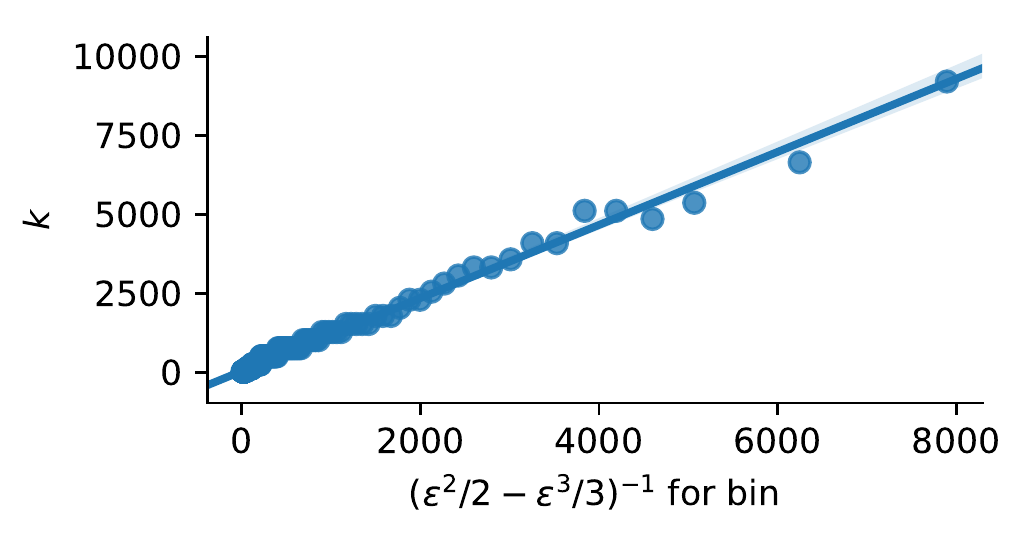}
    \caption{\small Minimum $\embsize$ sufficient for Rademacher embeddings to approximate BM25 pairwise rankings on \treccar with error rate ${\beta<.05}$.}
    \label{fig:simple-dp-approx-rankings}
\end{figure}

\paragraph{Pairwise accuracy.}
We use empirical data to test the applicability of Lemma 1 to the \bm{} relevance model. We select query-document triples $(q, d_1, d_2)$ from the \treccar dataset~\cite{treccar18} by considering all possible $(q, d_2)$, and selecting $d_1 = \text{arg}\max_{d}\text{\bm}(q, d)$. We bin the triples by the normalized margin $\epsilon$, and compute the quantity $(\epsilon^2/2 - \epsilon^3/3)^{-1}$.  According to Lemma~\ref{lem:ranking-error-rp}, the minimum embedding size of a random projection $k^*$ which has $\leq \beta$ probability of making an error on a triple with normalized margin $\epsilon$ is upper bounded by a linear function of this quantity. In particular, for $\beta = .05$, the Lemma entails that $k^* \leq 8.76 (\epsilon^2/2 - \epsilon^3/3)^{-1}$.  In this experiment we measure the empirical value of $k^*$ to evaluate the tightness of the bound.

The results are shown on the $x$-axis of \autoref{fig:simple-dp-approx-rankings}. For each bin we compute the minimum embedding size required to obtain 95\% pairwise accuracy in ranking $d_1$ vs $d_2$, using a grid of $40$ possible values for $\embsize$ between 32 and 9472, shown on the $y$-axis. (We exclude examples that had higher values of $(\epsilon^2/2 - \epsilon^3/3)^{-1}$ than the range shown because they did not reach 95\% accuracy for the explored range of $\embsize$.)  The figure shows that the theoretical bound is tight up to a constant factor, and that the minimum embedding size that yields desired fidelity grows linearly with $(\epsilon^2/2 - \epsilon^3/3)^{-1}$.  

\begin{figure*}
\centering
\subcaptionbox{Each datapoint is the median normalized margin per bin, and the shaded areas show the 25th and 75th quantiles.    \label{fig:bm25-margins}}
{    
\includegraphics[width=.44\linewidth]{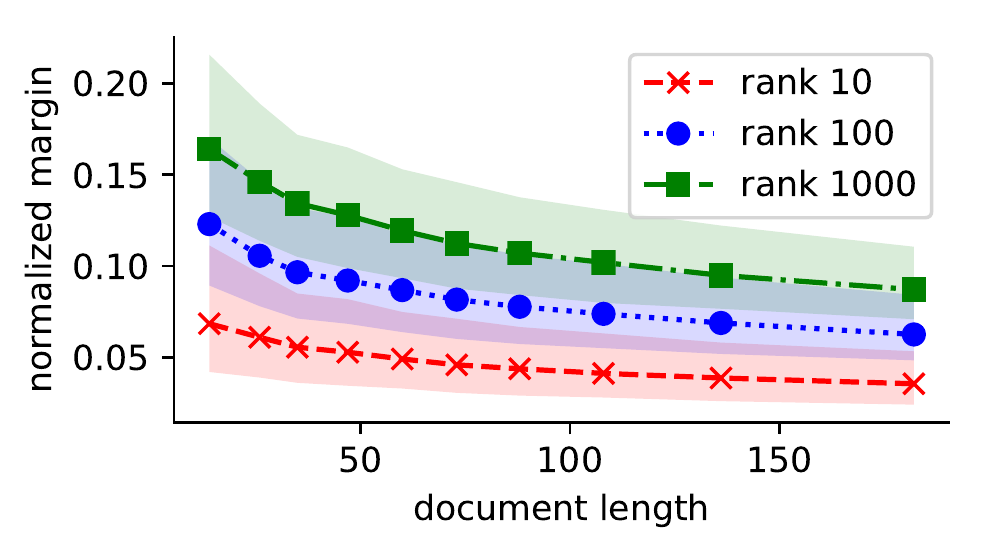}
}
\quad
\subcaptionbox
{Each line shows the minimum random projection dimension $\embsize$ that achieves a desired value of recall-at-$10$ for each bin of documents.
\label{fig:bm25-r10}
}
{   
\includegraphics[width=.44\linewidth]{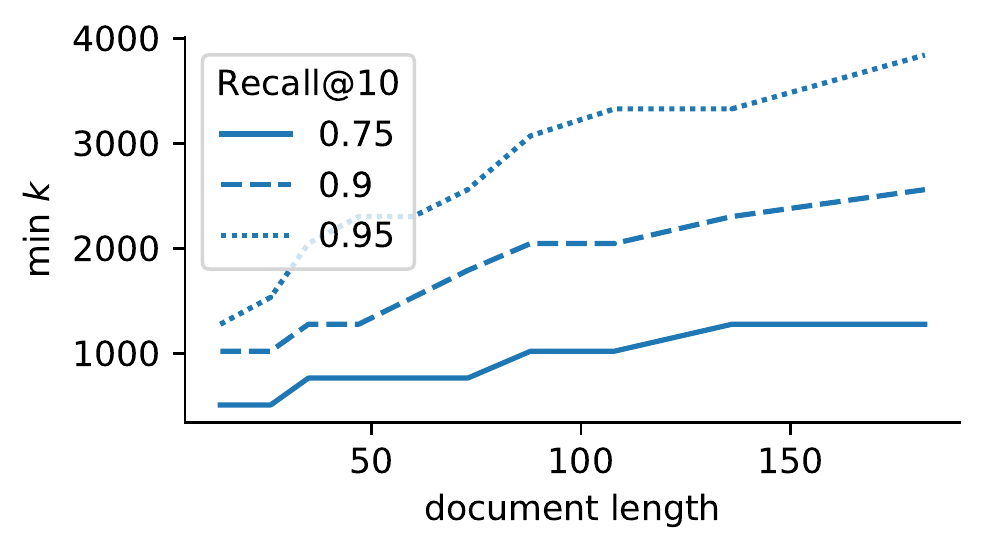}
}
\caption{Random projection on \bm retrieval in \treccar dataset, with documents binned by length.}\label{fig:bm25-trec}
\end{figure*}

\paragraph{Margins and document length.} 
For boolean inner product, it was possible to express the minimum possible normalized margin (and therefore a sufficient embedding size) in terms of $L_Q$ and $L_D$, the maximum number of unique terms across all queries and documents, respectively.  
Unfortunately, it is difficult to analytically derive a minimum normalized margin $\epsilon$ for either \tfidf or \bm: because each term may have a unique inverse document frequency, the minimum non-zero margin $\iprod{q}{d_1 - d_2}$ decreases with the number of terms in the query as each additional term creates more ways in which two documents can receive nearly the same score. We therefore study empirically how normalized margins vary with maximum document length. Using the \treccar retrieval dataset, we bin documents by length. For each query, we compute the normalized margins between the document with best \bm in the bin and all other documents in the bin, and look at the 10th, 100th, and 1000th smallest normalized margins. The distribution over these normalized margins is shown in \autoref{fig:bm25-margins}, revealing that normalized margins decrease with document length. In practice, the observed minimum normalized margin for a collection of documents and queries is found to be much lower for \bm compared to boolean inner product. For example, for the collection used in \autoref{fig:simple-dp-approx-rankings}, the minimum normalized margin for \bm is 6.8e-06, while for boolean inner product it is $0.0169$.

\paragraph{Document length and encoding dimension.}  \autoref{fig:bm25-r10} shows the growth in minimum random projection dimension required to reach desired recall-at-$10$, using the same document bins as in \autoref{fig:bm25-margins}. As predicted, the required dimension increases with the document length, while the  normalized margin shrinks.

\subsection{Bounds on general encoding functions}
\label{sec:general-limits}
We derived upper bounds on minimum required encoding for random linear projections above, and found the bounds on $(q,d_1,d_2)$ triples  to be empirically tight up to a constant factor. More general non-linear and learned encoders could be more efficient. However, there are general theoretical results showing that it is impossible for any encoder to guarantee an inner product distortion ${|\iprod{f(x)}{f(y)} - \iprod{x}{y}| \leq \epsilon}$ with an encoding that does not grow as $\Omega(\epsilon^{-2})$~\cite{larsen2017optimality,alon2017optimal}, for vectors $x,y$ with norm $\leq 1$. 
These results suggest more general capacity limitations for fixed-length dual encoders when document length grows.

In our setting, \bm, \tfidf, and boolean inner product can all be reformulated equivalently as inner products in a space with vectors of norm at most 1 by $L_2$-normalizing each query vector and rescaling all document vectors by $\sqrt{L_D} = \max_d ||d||$, a constant factor that grows with the length of the longest document.
Now suppose we desire to limit the distortion on the \emph{unnormalized} inner products to some value $\leq \tilde{\epsilon}$, which might guarantee a desired performance characteristic.
This corresponds to decreasing the maximum \emph{normalized} inner product distortion $\epsilon$ by a factor of $\sqrt{L_D}$. \newtextfinal{According to the general bounds on dimensionality reduction mentioned in the previous paragraph, this could necessitate an increase in the encoding size by a factor of $L_D$.}

However, there a number of caveats to this theoretical argument. First, the theory states only that there \emph{exist} vector sets that cannot be encoded into representations that grow more slowly than $\Omega(\epsilon^{-2})$; actual documents and queries might be easier to encode if, for example, they are generated from some simple underlying stochastic process. 
Second, our construction achieves $||d||\leq 1$ by rescaling all document vectors by a constant factor, but there may be other ways to constrain the norms while using the embedding space more efficiently.
\newtextfinal{Third, in the non-linear case it might be possible to eliminate ranking errors without achieving low inner product distortion.}
Finally, from a practical perspective, the generalization offered by learned dual encoders might overwhelm any sacrifices in fidelity, when evaluated on real tasks of interest. Lacking theoretical tools to settle these questions, we present a set of empirical investigations in later sections of this paper. 
But first we explore a lightweight modification to the dual encoder, which offers gains in expressivity at limited additional computational cost. 

%% file: multivector.tex
\section{Multi-Vector Encodings}
\label{sec:multivector}
The theoretical analysis suggests that fixed-length vector representations of documents may \newtext{in general need to be large for long documents, if fidelity with respect to sparse high-dimensional representations is important}. Cross-attentional architectures can achieve higher fidelity, \newtextfinal{but are impractical for large-scale retrieval ~\cite{doc-expansion,reimers-gurevych-2019-sentence,humeau-etal-2020}.}
We therefore propose a new architecture that represents each document as a fixed-size set of $m$ vectors. Relevance scores are computed as the maximum inner product over this set.

Formally, let $\xtoks = (x_1, \ldots, x_{\maxtoks})$ represent a sequence of tokens, with $x_1$ equal to the special token \clstok, and define $\ytoks$ analogously. Then $[h_1(\xtoks), \ldots, h_{\maxtoks}(\xtoks)]$ represents the sequence of contextualized embeddings at the top level of a deep transformer. We define a single-vector representation of the query $\xtoks$ as $f^{(1)}(\xtoks)=h_1(\xtoks)$, and a multi-vector representation of document $\ytoks$ as $f^{(m)}(\ytoks) = [h_1(\ytoks),\ldots,h_m(\ytoks)]$, the first $m$ representation vectors for the sequence of tokens in $\ytoks$, with $m < \maxtoks$.
The relevance score is defined as $\max_{j=1\ldots m} \iprod{f^{(1)}(\xtoks)}{f^{(m)}_j(\ytoks)}$.

Although this scoring function is not a dual encoder, the search for the highest-scoring document can be implemented efficiently with standard approximate nearest-neighbor search by adding multiple ($m$) entries for each document to the search index data structure. If some vector $f^{(m)}_j(\ytoks)$ yields the largest inner product with the query vector $f^{(1)}(\xtoks)$, it is easy to show the corresponding document must be the one that maximizes the relevance score $\psi^{(m)}(\xtoks, \ytoks).$  The size of the index must grow by a factor of $m$, but due to the efficiency of contemporary approximate nearest neighbor and maximum inner product search, the time complexity can be sublinear in the size of the index~\cite{ANDONI_2019,MIPS}. Thus, a model using $m$ vectors of size $k$ to represent documents is more efficient at run-time than a dual encoder that uses a single vector of size $m k$. 

This efficiency is a key difference from the {\textsc{Poly-encoder}}~\cite{humeau-etal-2020}, which computes a fixed number of vectors per \emph{query}, and aggregates them by softmax attention against document vectors. \newcite{bottleneck} propose a similar architecture for language modeling. Because of the use of softmax in these approaches, it is not possible to decompose the relevance score into a max over inner products, and so fast nearest-neighbor search cannot be applied. \newtext{In addition, these works did not address retrieval from a large document collection.}

\paragraph{Analysis.} To see why multi-vector encodings can enable smaller encodings per vector,
consider an idealized setting in which each document vector is the sum of $m$ orthogonal \term{segments} such that $d = \sum_{i=1}^m d^{(i)}$ and each query refers to exactly one segment in the gold document.\footnote{Here we use $(d, q)$ rather than $(\xtoks, \ytoks)$ because we describe vector encodings rather than token sequences.}
\newtextfinal{An orthogonal segmentation can be obtained by choosing the segments as a partition of the vocabulary.}
\begin{theorem}
\label{thm:multivector}
Define vectors $q, d_1, d_2 \in \reals^{\vocsize}$ such that $\iprod{q}{d_1} > \iprod{q}{d_2}$, and assume that both $d_1$ and $d_2$ can be decomposed into $m$ segments such that: $d_1 = \sum_{i=1}^m d_1^{(i)}$, and analogously for $d_2$; all segments across both documents are orthogonal. If there exists an $i$ such that $\iprod{q}{d_1} = \iprod{q}{d_1^{(i)}}$ and $\iprod{q}{d_2} \geq \iprod{q}{d_2^{(i)}}$,
then $\nmarg(q, d_1^{(i)}, d_2^{(i)}) \geq \nmarg(q, d_1, d_2)$.
(The proof is in \autoref{sec:proof-multivector}.)
\end{theorem}

\begin{remark}
The BM25 score can be computed from non-negative representations of the document and query; if the segmentation corresponds to a partition of the vocabulary, then the segments will also be non-negative, and thus the condition $\iprod{q}{d_2} \geq \iprod{q}{d_2^{(i)}}$ holds for all $i$.
\end{remark}

\newtextfinal{
The relevant case is when the same segment is maximal for both documents, $\iprod{q}{d_2^{(i)}} = \max_{j} \iprod{q}{d^{(j)}_2}$, as will hold for ``simple'' queries that are well-aligned with the segmentation. Then the normalized margin in the multi-vector model will be at least as large as in the equivalent single vector representation. The relationship to encoding size follows from the theory in the previous section: \autoref{thm:multivector} implies that if we set $f^{(m)}_i(\ytoks) = A d^{(i)}$ (for appropriate $A$), then an increase in the normalized margin enables the use of a smaller encoding dimension $\embsize$ while still supporting the same pairwise error rate. There are now $m$ times more ``documents'' to evaluate, but Lemma~\ref{lem:recall-at-r} shows that this exerts only a logarithmic increase on the encoding size for a desired recall@$r$. But while we hope this argument is illuminating, the assumptions of orthogonal segments and perfect segment match against the query are quite strong.
We must therefore rely on empirical analysis to validate the efficacy of multi-vector encoding in realistic applications.}

\newtext{\paragraph{Cross-attention.} Cross-attentional architectures can be viewed as a generalization of the multi-vector model: (1) set $m=T_{\text{max}}$ (one vector per token); (2) compute one vector per token in the query; (3) allow more expressive aggregation over vectors than the simple $\max$ employed above. Any sparse scoring function (e.g., \bm) can be mimicked by a cross-attention model, which need only compute identity between individual words; this can be achieved by random projection word embeddings whose dimension is proportional to the log of the vocabulary size. By definition, the required representation also grows linearly with the number of tokens in the passage and query. As with the  {\textsc{Poly-encoder}}, retrieval in the cross-attention model cannot be performed efficiently at scale using fast nearest-neighbor search.}  In contemporaneous work,~\newcite{colbert} propose an approach with $T_Y$ vectors per query and $T_X$ vectors per document, using a simple sum-of-max for aggregation of the inner products.  They apply this approach to retrieval via re-ranking results of $T_Y$ nearest-neighbor searches. Our multi-vector model uses fixed length representations instead, and a single nearest neighbor search per query.

%% file: experiments.tex
\section{Experimental Setup}
\label{sec:experiments}

\newtextbresub{The full IR task requires detection of both precise word overlap and semantic generalization.  Our theoretical results focus on the first aspect, and derive theoretical and empirical bounds on the sufficient dimensionality to achieve high fidelity with respect to sparse bag-of-words models as document length grows, for  two types of linear random projections.} The theoretical setup differs from modeling for realistic information-seeking scenarios in at least two ways.



\newtextbresub{First, trained non-linear dual encoders might be able to detect precise word overlap with much lower-dimensional encodings, especially for queries and documents with a natural distribution, which may exhibit a low-dimensional subspace structure.  Second, the  semantic generalization aspect of the IR task may be more important than the first aspect for practical applications, and our theory does not make predictions about how encoder dimensionality  relates to such ability to compute general semantic similarity.}




We relate the theoretical analysis to text retrieval in practice through experimental studies on three tasks. The first task, described in \autoref{sec:ict-unmasked}, tests the ability of models to retrieve natural language documents that exactly contain a query and evaluates both \bm and deep neural dual encoders on a task of detecting precise word overlap, defined over texts with a natural distribution. The second task, described in \autoref{sec:nq}, is the passage retrieval sub-problem of the open-domain QA version of the Natural Questions \cite{nq,lee-etal-2019-latent}; this benchmark reflects the need to capture graded notions of similarly and has a natural query text distribution. For both of these tasks, we perform controlled experiments varying the maximum length of the documents in the collection, which enables assessing the relationship between encoder dimension and document length. 

To evaluate the performance of our best models in comparison to state-of-the-art works on large-scale retrieval and ranking, in ~\autoref{sec:ir-eval} we report results on a third group of tasks focusing on passage/document ranking: the passage and document-level {\msmarco} retrieval datasets~\cite{nguyen2016ms,craswell2020overview}. Here we follow the standard two-stage retrieval and ranking system: a first-stage retrieval from a large document collection, followed by reranking with a cross-attention model. We focus on the impact of the first-stage retrieval model. 

\subsection{Models}
\label{sec:models}

Our experiments compare compressive and sparse dual encoders, cross attention, and hybrid models.


\paragraph{{\bm}.}
We use case-insensitive wordpiece tokenizations of texts
and default \bm parameters from the \textit{gensim} library. We apply either unigram (\bmone) or combined unigram+bigram representations (\bmtwo). 

\paragraph{Dual encoders from \bert ({\smaller{DE}-\bert}).}
We encode queries and documents using \bert-base, which is a pre-trained transformer network  (12 layers, 768 dimensions)~\cite{devlin-etal-2019-bert}. We implement dual encoders from \bert as a special case of the multi-vector model formalized in \autoref{sec:multivector}, with number of vectors for the document $m=1$: the representations for queries and documents are the top layer representations at the \clstok token.  This approach is widely used for retrieval~\cite{lee-etal-2019-latent,reimers-gurevych-2019-sentence, humeau-etal-2020,xiong2020approximate}.\footnote{ Based on preliminary experiments with pooling strategies  we use the \clstok vectors (without the feed-forward projection learned on the next sentence prediction task).}
For lower-dimensional encodings, we learn down-projections from $d=768$ to $\embsize \in$ $ 32,64, 128, 512$,\footnote{We experimented with adding a similar layer for $d=768$, but this did not offer empirical gains.} implemented as a single feed-forward  layer, followed by layer normalization. All parameters  are  fine-tuned for the retrieval tasks. We refer to these models as {\de-$\embsize$}.

\paragraph{Cross-Attentional \bert.}
The most expressive model  we consider is cross-attentional \bert, which we implement by applying the \bert encoder to the concatenation of the query and document, with a special \textsc{[sep]} separator  between $\xtoks$ and $\ytoks$.
The relevance score is  a learned linear function of the encoding of the \clstok token.
Due to the computational cost, cross-attentional \bert is applied only in reranking as in prior work \cite{passage-rerank, bert-adhoc}. These models are referred to as {\crossatt}.
\commentout{\paragraph{Sum-of-Max}
As a more lightweight alternative to  {\crossatt}, we compute a score by separately encoding the query $x$ and document $y$, and summing over the maximum inner product for each token in the query, 
$\sum_{t=1}^{\maxtoks_x}\max_{t' \in [\maxtoks_y]} \iprod{x_t}{y_{t'}},$ where $x_t$ is the \bert contextualized embedding for token $t$ in the query and $y_{t'}$ is the contextualized embedding for token $t'$ in the document. This model is closely related to the ``hard attention'' model that was analyzed in \autoref{sec:attention}. Although it cannot be efficiently implemented at large scale, sum-of-max is considerably faster than cross attentional \bert . Considering prediction-time cost, sum-of-max is only $O(\embsize \maxtoks_x \maxtoks_y ),$ whereas {\crossatt} needs to jointly encode $x$ and $y$, leading to a cost of $O(\embsize (\maxtoks_x+\maxtoks_y)^2 )$ per layer~\cite{vaswani2017attention}. }

\paragraph{Multi-Vector Encoding from BERT ({\smaller{ME}-\bert}).}
In \autoref{sec:multivector} we introduced a model in which every document is represented by exactly $m$ vectors. We use $m=8$ as a good compromise between cost and accuracy in  \autoref{sec:ict-unmasked} and \autoref{sec:nq}, and find values of $3$ to $4$ for $m$ more accurate on the datasets in~\autoref{sec:ir-eval}. 
In addition to using \bert output representations directly, we  consider down-projected representations, implemented using  a  feed-forward layer with dimension $768 \times k$.  \commentout{Unlike for the \de-\bert models, for the multi-vector case we found that an additional projection layer helps even when $k=768$.}
A  model with $k$-dimensional embeddings is referred to as {\smaller{ME}-\bert-$k$}. 


\paragraph{Sparse-Dense Hybrids (\textsc{\small{Hybrid}}).}
A natural approach to balancing between the fidelity of sparse representations  and the generalization of learned dense ones is to build a hybrid. To do this, we linearly combine a sparse and dense system's scores using  a single trainable weight $\lambda$, tuned on a development set. For example, a hybrid model of {\smaller{ME}-{\bert}} and {\bmone} is referred to as 
{\small{\hybrid}-{\qonedeight}-\small{uni}}. We implement approximate search to retrieve using a linear combination of two systems by re-ranking $n$-best top scoring candidates from each system. 
Prior  and concurrent work has also used hybrid sparse-dense models ~\cite{guo16,seo-etal-2019-real,karpukhin2020dense,ma2020zeroshot,gao2020complementing}. Our contribution is to assess the impact of  sparse-dense hybrids as the document length grows.


\subsection{Learning and Inference}

For the experiments in ~\autoref{sec:ict-unmasked} and ~\autoref{sec:nq}, all trained models are initialized from  \bert-base, and all parameters are fine-tuned using a cross-entropy loss with $7$ sampled negatives from a pre-computed $200$-document list and additional in-batch negatives (with a total number of 1024 candidates in a batch); the pre-computed candidates include $100$ top neighbors from \bm and  $100$ random samples.  This is similar to the method by \newcite{lee-etal-2019-latent}, but with additional fixed candidates, also used in concurrent work~\cite{karpukhin2020dense}. Given a model trained in this way, for the scalable methods, we also applied hard-negative mining as in \newcite{gillick-etal-2019-learning} and used one iteration  when beneficial. More sophisticated negative selection is proposed in concurrent work ~\cite{xiong2020approximate}. For retrieval from large document collections with the scalable models, we used ScaNN: an efficient approximate nearest neighbor search library~\cite{scann}; in most experiments, we use exact search settings but also evaluate approximate search in Section ~\autoref{sec:ir-eval}.  In ~\autoref{sec:ir-eval}, the same general approach with slightly different hyperparameters (detailed in that section) was used, to enable more direct comparisons to prior work.

\section{Containing Passage ICT Task}
\label{sec:ict-unmasked}
\input{figures/unmasked_figure_new_3.tex}




We begin with experiments on the task of retrieving a Wikipedia passage $y$ containing a sequence of words $x$. 
We create a dataset using Wikipedia, following the Inverse Cloze Task definition by \newcite{lee-etal-2019-latent}, but adapted to suit the goals of our study. The task is defined by first breaking Wikipedia texts into segments  of length at most $l$. These form the document collection ${\cal{D}}$. Queries $x_i$ are generated by sampling sub-sequences from the documents $y_i$.  We use queries of lengths between $5$ and $25$, and do not remove the queries $x_i$ from their corresponding documents $y_i$.  

We create a dataset with one million queries and evaluate retrieval against four document collections ${\cal{D}}_l$, for $l \in $ $50,100,200,400$. Each  ${\cal{D}}_l$ contains three million documents of maximum length $l$ tokens.  In addition to original Wikipedia passages, each ${\cal{D}}_l$ contains synthetic distractor documents, which  contain the large majority of words in $x$ but differ by one or two tokens.  5K queries are used for evaluation, leaving the rest for training and validation. Although checking containment is  a straightforward machine learning task,  it is a good testbed for assessing the fidelity of compressive neural models.  \bmtwo achieves over 95 MRR@10  across collections for this task. 


\autoref{fig:unmasked} (\textit{left}) shows test set results on  reranking, where models need to select one of 200 passages (top 100 \bmtwo and 100 random candidates). It is interesting to see how strong the sparse  models are relative to  even a  768-dimensional \de. As the document length increases, the performance of both the sparse and dense dual encoders worsens; the accuracy of the \de models falls most rapidly, widening the gap to \bm.

Full cross attention is nearly perfect and does not degrade with document length.   {\qonedeight-{\smaller{768}}} which uses 8 vectors of dimension 768 to represent documents strongly outperforms the best \de model. Even \qonedeight-{\smaller{64}}, which uses 8 vectors of size only 64 instead (thus requiring the same document collection size as \de-{\smaller{512}} and being faster at inference time), outperforms the \de models by a large margin.
\commentout{It was not feasible to evaluate \de models with larger embedding size, but Rademacher embeddings were observed to require  $\embsize$ of approximately  4K, 6K, 8K, and 32K, for the four document collections, respectively, to achieve 99\% of \bmone's accuracy. }

\autoref{fig:unmasked} (\textit{right}) shows results for the much more challenging task of retrieval from three million candidates. For the latter setting, we only evaluate models that can efficiently retrieve nearest neighbors from such a large set. We see similar behavior to the reranking setting, with the multi-vector methods exceeding \bmone performance for all lengths and \de models under-performing \bmone. The hybrid model outperforms both components in the combination with largest improvements over \qonedeight for the longest-document collection.







\section{Retrieval for Open-domain QA}
\label{sec:nq}
\input{figures/nq_figures_new.tex}

For this task we similarly use English Wikipedia\footnote{\url{https://archive.org/download/enwiki-20181220}} as four different document collections, of maximum passage length $l \in \{50, 100, 200, 400\}$, and corresponding approximate sizes of 39 million, 27.3 million, 16.1 million, and 10.2 million documents, respectively. Here we use real user queries contained in the Natural Questions dataset~\cite{nq}. We follow the setup in \newcite{lee-etal-2019-latent}. There are $87,925$ QA pairs in training and $3,610$ QA pairs in the test set. We hold out a subset of training for development. 

For document retrieval, a passage is correct for a query $x$ if it contains a string that matches exactly an annotator-provided short answer for the question. We  form a reranking task by considering the top $100$ results from \bmone and  $100$ random samples, and also consider the full retrieval setting. \bmone is used here instead of \bmtwo, because it is the stronger model for this task.

Our theoretical results do not make direct predictions for performance of compressive dual encoder models relative to \bm on this task.  They do tell us that as the document length grows, low-dimensional compressive dual encoders may not be able to measure weighted term overlap precisely, potentially leading to lower performance on the task. Therefore, we would expect that  higher dimensional dual encoders, multi-vector encoders, and hybrid models become more useful for collections with longer documents.

\autoref{fig:nq} (\textit{left}) shows heldout set results on the reranking task. To fairly compare systems that operate over  collections of different-sized passages, we allow each model to select approximately the same number of tokens ($400$) and evaluate on whether an answer is contained in them. For example, models retrieving from ${\cal D}_{50}$ return their top 8 passages, and ones retrieving from ${\cal D}_{100}$ retrieve top 4. The figure shows this recall@400 tokens across models. The relative performance of \bmone and  \de  is different from that seen in the ICT  task, due to the semantic generalizations needed. Nevertheless, higher-dimensional \de models generally perform better, and multi-vector  models provide further benefits, especially for longer-document collections;   \qonedeight-{\smaller{768}} outperforms \de-{\smaller{768}} and  \qonedeight-{\smaller{64}} outperforms \de-{\smaller{512}}; \crossatt{ } is still substantially stronger. 

\autoref{fig:nq} (\textit{right}) shows heldout set results  for the  task of retrieving from  Wikipedia for each of the four document collections ${\cal{D}}_l$. Unlike the reranking setting, only higher-dimensional  \de models outperform \bm  for passages longer than 50.  The hybrid models offer large improvements over their components, capturing both precise word overlap and  semantic similarity. The gain from adding \bm to \qonedeight and \de increases as the length of the documents in the collection grows, which is consistent with our expectations based on the theory.

%% file: figures/unmasked_figure_new_3.tex
\begin{figure*}

\begin{tikzpicture}
    \pgfplotsset{footnotesize,samples=10}
    \begin{groupplot}[group style = {group size = 2 by 1, horizontal sep = 50pt}, width = 6.5cm, height = 4cm]
    \nextgroupplot[
            xticklabels from table={passage_sizes.dat}{Label},
            xticklabel style={align=center,font=\small},
            legend style = { column sep = 10pt, legend columns = -1, legend to name = grouplegend,},
            legend columns=2,
            every tick label/.append style={font=\small},
            legend style={column sep=0.2cm},
            legend style={anchor=south},
            legend style={at={(0.5, 0.0), anchor=south}, font=\large},
            ymin=45,ymax=98,
            enlarge y limits=true,
            title={{Passage Ranking for ICT}}, ylabel={MRR@10},
            ]
             \addrecallplotdethreetwo{figures/data/results/unmasked_rerank_efficient_test_MRR.csv};\label{addrecallplotdethreetwo}
                         \addrecallplotdesixfour{figures/data/results/unmasked_rerank_efficient_test_MRR.csv};\label{addrecallplotdesixfour}
                \addrecallplotdeonetwoeight{figures/data/results/unmasked_rerank_efficient_test_MRR.csv};\label{addrecallplotdeonetwoeight}
                \addrecallplotdefivetwelve{figures/data/results/unmasked_rerank_efficient_test_MRR.csv};\label{addrecallplotdefivetwelve}
                \addrecallplotdesevensixeight{figures/data/results/unmasked_rerank_efficient_test_MRR.csv};
    \addrecallplotbmone{figures/data/results/unmasked_rerank_efficient_test_MRR.csv};
        \addrecallplotbmtwo{figures/data/results/unmasked_rerank_efficient_test_MRR.csv};
    \addrecallplotqonedeight{figures/data/results/unmasked_rerank_efficient_test_MRR.csv};\label{addrecallplotqonedeight}

            \addrecallplotqonedeightsixfour{figures/data/results/unmasked_rerank_efficient_test_MRR.csv};\label{addrecallplotqonedeightsixfour}

              \addrecallplotca{figures/data/results/unmasked_rerank_efficient_test_MRR.csv};\label{addrecallplotca}
             

\coordinate (top) at (rel axis cs:0,1);
    \nextgroupplot[
            xticklabels from table={passage_sizes.dat}{Label},
            xticklabel style={align=center,font=\small},
            every tick label/.append style={font=\small},
            ymin=30,ymax=98,
            enlarge y limits=true,
            title={{Passage Retrieval for  ICT}}, ylabel={MRR@10},
            ]
             \addrecallplotdethreetwo{figures/data/results/unmasked_retrieve_efficient_test_MRR.csv};
             \addrecallplotdesixfour{figures/data/results/unmasked_retrieve_efficient_test_MRR.csv};
            \addrecallplotdeonetwoeight{figures/data/results/unmasked_retrieve_efficient_test_MRR.csv};
            \addrecallplotdefivetwelve{figures/data/results/unmasked_retrieve_efficient_test_MRR.csv};
            \addrecallplotdesevensixeight{figures/data/results/unmasked_retrieve_efficient_test_MRR.csv};
            \addrecallplotqonedeight{figures/data/results/unmasked_retrieve_efficient_test_MRR.csv};
           \addrecallplotqonedeightsixfour{figures/data/results/unmasked_retrieve_efficient_test_MRR.csv};
    \addrecallplotbmone{figures/data/results/unmasked_retrieve_efficient_test_MRR.csv};
        \addrecallplotbmtwo{figures/data/results/unmasked_retrieve_efficient_test_MRR.csv};
            
            \addrecallplotbmqonedeightfull{figures/data/results/unmasked_retrieve_efficient_test_MRR.csv};\label{addrecallplotbmqonedeightfull}
            
            
           \addrecallplotbmqonedeightfullbi{figures/data/results/unmasked_retrieve_efficient_test_MRR.csv};\label{addrecallplotbmqonedeightfullbi}
            
\coordinate (bot) at (rel axis cs:1,0);
    \end{groupplot}
    
    \path (top)--(bot) coordinate[midway] (group center);
        
  \matrix[
      matrix of nodes,
      anchor=west,
      inner sep=0.13em,
    ]at([right=1em,inner sep=0pt]group center -| current bounding box.east)
    { \ref{addrecallplotca}&\small{\crossatt}  \\
      \ref{addrecallplotdethreetwo}& \small{\de}-\small{32}  \\
            \ref{addrecallplotdesixfour}& \small{\de}-\footnotesize{64}  \\
            \ref{addrecallplotdeonetwoeight}& \small{\de}-\footnotesize{128}  \\
                        \ref{addrecallplotdefivetwelve}& \small{\de}-\footnotesize{512}  \\
                 \ref{addrecallplotdesevensixeight}& \small{\de}-\footnotesize{768}  \\
                                        \ref{addrecallplotqonedeightsixfour}& \small{\qonedeight}-\footnotesize{64} \\

                       \ref{addrecallplotqonedeight}& \small{\qonedeight}-\footnotesize{768} \\

            \ref{addrecallplotbmqonedeightfull}& \footnotesize{\hybrid}-{\qonedeight}-\footnotesize{uni} \\
            \ref{addrecallplotbmqonedeightfullbi}& \footnotesize{\hybrid}-{\qonedeight}-\footnotesize{bi} \\
            
            \ref{addrecallplotbmone}& \small{\bmone} \\
            \ref{addrecallplotbmtwo}& \small{\bmtwo} \\
};

\end{tikzpicture}
                \caption{\small Results on the containing passage ICT task as maximum passage length varies (50 to 400 tokens). \textit{Left}: Reranking 200 candidates; \textit{Right}: Retrieval  from  three  million  candidates. Exact numbers refer to Table \ref{tab:ICT_NQ}. 
                }
                \label{fig:unmasked}

\end{figure*}
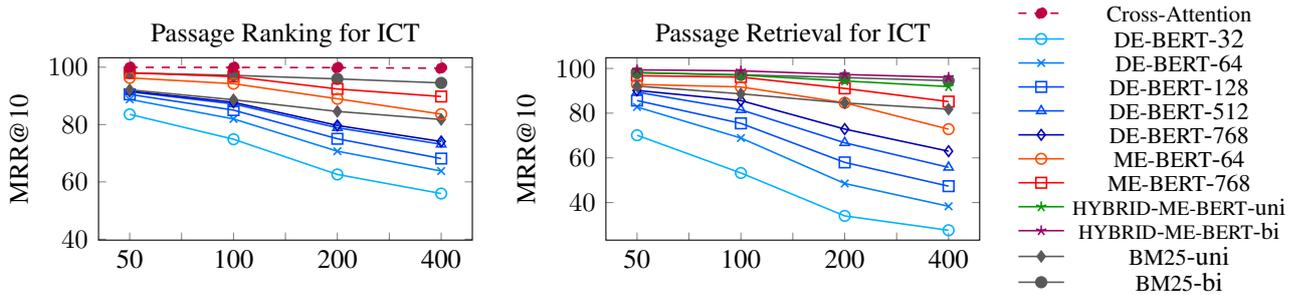

%% file: figures/nq_figures_new.tex
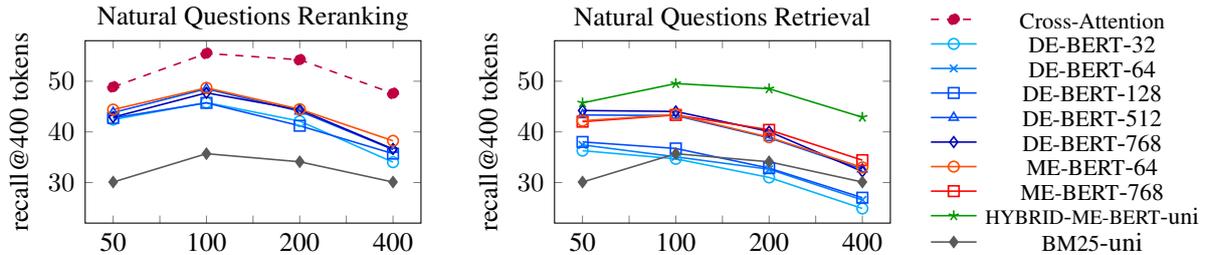
\begin{figure*}

\begin{tikzpicture}
    \pgfplotsset{footnotesize,samples=10}
    \begin{groupplot}[group style = {group size = 2 by 1, horizontal sep = 50pt}, width = 6.0cm, height = 4cm]
    \nextgroupplot[
            xticklabels from table={passage_sizes.dat}{Label},
            xticklabel style={align=center,font=\small},
            legend style = { column sep = 10pt, legend columns = -1, legend to name = grouplegend,},
            legend columns=2,
            every tick label/.append style={font=\small},
            legend style={column sep=0.2cm},
            legend style={anchor=south},
            legend style={at={(0.5, 0.0), anchor=south}, font=\large},
            ymin=25,ymax=55,
            enlarge y limits=true,
            title={{Natural Questions Reranking}}, ylabel={recall@400 tokens},
            ]
            \addrecallplotdethreetwo{figures/data/results/NQ_rerank_efficient_dev.csv};
                \addrecallplotdeonetwoeight{figures/data/results/NQ_rerank_efficient_dev.csv};
                \addrecallplotdefivetwelve{figures/data/results/NQ_rerank_efficient_dev.csv};
              \addrecallplotdesevensixeight{figures/data/results/NQ_rerank_efficient_dev.csv};
 \addrecallplotbmone{figures/NQ_bmone.csv};                  

            \addrecallplotqonedeightsixfour{figures/data/results/NQ_rerank_efficient_dev.csv};
           
              \addrecallplotca{figures/data/results/NQ_rerank_efficient_dev.csv};
             

\coordinate (top) at (rel axis cs:0,1);
 \nextgroupplot[
            xticklabels from table={passage_sizes.dat}{Label},
            xticklabel style={align=center,font=\small},
            every tick label/.append style={font=\small},
            ymin=25,ymax=55,
            enlarge y limits=true,
            title={{Natural Questions Retrieval}}, ylabel={recall@400 tokens},
            ]
            \addrecallplotdethreetwo{figures/data/results/NQ_retrieve_efficient_dev.csv};
           \addrecallplotdesixfour{figures/data/results/NQ_retrieve_efficient_dev.csv};
             \addrecallplotdeonetwoeight{figures/data/results/NQ_retrieve_efficient_dev.csv};
            \addrecallplotdefivetwelve{figures/data/results/NQ_retrieve_efficient_dev.csv};
          \addrecallplotdesevensixeight{figures/data/results/NQ_retrieve_efficient_dev.csv};
           \addrecallplotqonedeight{figures/data/results/NQ_retrieve_efficient_dev.csv};
           \addrecallplotqonedeightsixfour{figures/data/results/NQ_retrieve_efficient_dev.csv};
            
           \addrecallplotbmqonedeightfull{figures/data/results/NQ_retrieve_efficient_dev.csv};\label{addrecallplotbmqonedeightfull}
            \addrecallplotbmone{figures/NQ_bmone.csv};
\coordinate (bot) at (rel axis cs:1,0);

    \end{groupplot}
    
    \path (top)--(bot) coordinate[midway] (group center);
        
  \matrix[
      matrix of nodes,
      anchor=west,
      inner sep=0.14em,
    ]at([right=1em,inner sep=0pt]group center -| current bounding box.east)
    { \ref{addrecallplotca}&\small{\crossatt}  \\
      \ref{addrecallplotdethreetwo}& \small{\de}-\footnotesize{32}  \\
            \ref{addrecallplotdesixfour}& \small{\de}-\footnotesize{64}  \\
            \ref{addrecallplotdeonetwoeight}& \small{\de}-\footnotesize{128}  \\
                        \ref{addrecallplotdefivetwelve}& \small{\de}-\footnotesize{512}  \\
                 \ref{addrecallplotdesevensixeight}& \small{\de}-\footnotesize{768}  \\
                                        \ref{addrecallplotqonedeightsixfour}& \small{\qonedeight}-\footnotesize{64} \\
                       \ref{addrecallplotqonedeight}& \small{\qonedeight}-\footnotesize{768} \\
          \ref{addrecallplotbmqonedeightfull}&
            \footnotesize{\hybrid}-{\qonedeight}-\footnotesize{uni} \\

            \ref{addrecallplotbmone}& \small{\bmone} \\
};

\end{tikzpicture}
                \caption{\small Results on NQ passage recall as maximum passage length varies (50 to 400 tokens). \textit{Left}: Reranking of 200 passages; \textit{Right}: Open domain retrieval result on all of (English) Wikipedia. Exact numbers refer to Table \ref{tab:ICT_NQ}. }
                \label{fig:nq}
\end{figure*}

%% file: experiments-retrieval.tex
\section{Large-Scale Supervised IR} 
\label{sec:ir-eval}

The previous  experimental sections focused on understanding the relationship between compressive encoder representation dimensionality and document length.
Here we evaluate whether our newly proposed multi-vector retrieval model {\smaller{ME-BERT}}, its corresponding dual encoder baseline  {\smaller{DE-BERT}}, and sparse-dense hybrids compare favorably to state-of-the-art models for large-scale supervised retrieval and ranking on IR benchmarks. 

\paragraph{Datasets.} The \msmarco passage ranking task focuses on ranking passages from a collection of about 8.8 mln. About 532k queries paired with relevant passages are provided for  training. The \msmarco document ranking task is on ranking full documents instead. The full collection contains about 3 million documents and the training set has about 367 thousand queries. We report results on the passage and document  development sets, comprising 6,980 and 5,193 queries, respectively in \autoref{tab:summary}.  We  report  \msmarco and TREC DL 2019 ~\cite{craswell2020overview} test  results in \autoref{tab:maintest}.

\paragraph{Model Settings.} For \msmarco passage we apply  models on the provided passage collections. For \msmarco document, we follow \newcite{alibaba20} and break documents into a set of overlapping passages with length up to 482 tokens, each including the document URL and title. For each task, we train the models on that task's training data only. We initialize the retriever and reranker models with \bert-large. We train dense retrieval models on positive and negative candidates from the 1000-best list of \bm, additionally using one iteration of hard negative mining when beneficial.  For {\smaller{ME-BERT}}, we used $m=3$ for the passage and $m=4$ for the document task.



\input{tables/retrieval_result_summary.tex}
\input{figures/MSMARCO_at_depth_figure.tex}

\paragraph{Results.}
\autoref{tab:summary} comparatively evaluates our models on the dev sets of two tasks. The state of the art prior work  follows the two-stage retrieval and reranking approach, where an efficient first-stage system retrieves a (usually large) list of candidates from the document collection, and a second stage more expensive model such as cross-attention \bert reranks the candidates.

Our focus is on improving the first  stage, and we compare to prior works in two settings: {\textbf{Retrieval}}, top part of \autoref{tab:summary}, where only first-stage efficient retrieval systems are used and  {\textbf{Reranking}}, bottom part of the table, where more expensive second-stage models are employed to re-rank candidates. \autoref{fig:msmarco_depth} delves into the impact of the first-stage retrieval systems as the number of candidates the second stage reranker has access to  is substantially reduced, improving   efficiency.

We report results in comparison to the following systems: 1) \textsc{Multi-Stage}~\cite{nogueira2019doc2query}, which reranks \bm  candidates with a cascade of \bert models.  2) \textsc{Doc2Query}~\cite{doc-expansion} and \textsc{DocT5Query}~\cite{nogueira2019doc2query}, which use neural models to expand documents before indexing and scoring with sparse retrieval models. 3) \textsc{DeepCT}~\cite{dai2019context}, which learns to map \bert's contextualized text representations to
context-aware term weights.
4) \textsc{HDCT}~\cite{dai2020context}  uses a hierachical approach that combines passage-level term weights into document level term weights. 5) \textsc{IDST},  a two-stage cascade ranking pipeline by \newcite{alibaba20}, and  6) Leaderboard, which is the best  score on the \msmarco-passage leaderboard as of Sept 18, 2020.\footnote{ \url{https://microsoft.github.io/msmarco/}}

We also compare our models to both our own \bm implementation described in ~\autoref{sec:models}, and external publicly available sparse model implementations, denoted with \smaller{BM25-E}\xspace. For the passage task, \smaller{BM25-E}\xspace is the Anserini~\cite{anserini} system with default parameters. For the document task, \smaller{BM25-E}\xspace is the official  IndriQueryLikelihood baseline. We report on dense-sparse hybrids using both our own \bm, and the external sparse systems; the latter hybrids are indicated by a suffix \textsc{-e}.


Looking at the top part of \autoref{tab:summary}, we can see that our \de model already outperforms or is competitive with prior systems. The multi-vector model brings larger improvement on the dataset containing  longer documents (\msmarco document), and the sparse-dense hybrid models bring improvements over dense-only models on both datasets. According to a Wilcoxon signed rank test for statistical significance, all differences between \textsc{de-bert}, \textsc{me-bert}, \textsc{de-hybrid-e}, and \textsc{me-hybrid-e} are statistically significant on both development sets with $p$-value $<.0001$.

When a large number of candidates can be reranked, the impact of the first-stage  system decreases. In the bottom part of the table we see that our models are comparable to systems reranking \bm candidates.The accuracy of the first-stage  system is particularly important when the cost of reranking a large set of candidates is prohibitive. \autoref{fig:msmarco_depth} shows the performance of systems that rerank a smaller number of candidates. We see that,  when a very small number of candidates can be scored with expensive cross-attention models, the multi-vector {\smaller{ME}}-\bert and hybrid models achieve large improvements compared to prior systems on both \msmarco tasks.

\autoref{tab:maintest} shows test results for dense models,  external sparse model baselines, and hybrids of the two (without reranking). In addition to test set (eval) results on the \msmarco passage task, we report metrics on the manually annotated passage and document retrieval test set at  TREC DL 2019. We report the fraction of unrated items as Holes@10 following~\citet{xiong2020approximate}.


\paragraph{Time and space analysis}
Figure ~\ref{fig:timeanalysis} compares the running time/quality trade-off curves for \textsc{de-bert} and \textsc{me-bert} on the \msmarco passage task using the ScaNN~\cite{scann} library on a 160 Intel(R) Xeon(R) CPU @ 2.20GHz cores machine with 1.88TB memory. Both models use one vector of size $k=1024$ per query; \textsc{de-bert} uses one and \textsc{me-bert} uses 3 vectors of size $k=1024$ per document. The size of the document index for \textsc{de-bert} is $34.2$GB and the size of the index for \textsc{me-bert} is about 3 times larger.  The indexing time was 1.52h and 3.02h for \textsc{de-bert} and \textsc{me-bert}, respectively. The ScaNN configuration we use is num\_leaves=5000, and num\_leaves\_to\_search ranges from 25 to 2000 (from less to more exact search) and time per query is measured when using parallel inference on all 160 cores.
In the higher quality range of the curves, \textsc{me-bert} achieves substantially higher MRR than \textsc{de-bert} for the same inference time per query. 

\input{figures/time_analysis}



%% file: tables/retrieval_result_summary.tex
\begin{table}[]
\newcolumntype{Y}{>{\centering\arraybackslash}X}
\setlength{\tabcolsep}{.1em}
\scriptsize
\scalebox{.87}{
\begin{tabularx}{\columnwidth}{l l c *{2}{Y}}
\toprule
    & & \multicolumn{1}{c}{MS-Passage} & \multicolumn{1}{c}{MS-Doc} \\ 
    \cmidrule(lr){3-3} \cmidrule(lr){4-4} 
     & Model & MRR  & MRR \\
    \midrule
Retrieval & BM25   & 0.167  &  0.249 \\
 & \textsc{BM25-E} & 0.184 &  0.209 \\
 & \textsc{Doc2Query}   & 0.215  & -  \\
 & \textsc{docT5query}  &  0.278  & - \\
& \textsc{DeepCT}  & 0.243  & - \\
 & \textsc{hdct}  & - & 0.300 \\
& \textsc{de-bert}  & 0.302  & 0.288 \\
& \textsc{me-bert}  & 0.334  & 0.333 \\
& \textsc{de-hybrid}   & 0.304  & 0.313 \\
& \textsc{de-hybrid-e}   & 0.309  & 0.315 \\
& \textsc{me-hybrid}   & {0.338}  & \textbf{0.346} \\
& \textsc{me-hybrid-e}   & \textbf{0.343}  & {0.339} \\
    \midrule
Reranking & \textsc{Multi-Stage}   & 0.390 & - \\
 & \textsc{idst}     & 0.408 & - \\

& Leaderboard    & \textbf{0.439}  & - \\
& \textsc{de-bert}   & 0.391  & 0.339 \\
& \textsc{me-bert}    & 0.395  & \textbf{0.353} \\
& \textsc{me-hybrid}   & 0.394  & \textbf{0.353}\\

\bottomrule
\end{tabularx}
}
\caption{{\small Development set results on \msmarco-Passage (MS-Passage), \msmarco-Document (MS-Doc) showing MRR@10.}}
  \label{tab:summary}
\end{table}

\begin{table}[]
\newcolumntype{Y}{>{\centering\arraybackslash}X}
\setlength{\tabcolsep}{.1em}
\scriptsize
\scalebox{.87}{
\begin{tabularx}{\columnwidth}{ l c *{4}{Y}}
\toprule
     Model & MRR(MS) & RR & NDCG@10 & Holes@10 \\ 
    \midrule
     \multicolumn{5}{c}{Passage Retrieval}  \\
     \midrule
 BM25-Anserini & 0.186 & 0.825 & 0.506 & \textbf{0.000} \\
 \textsc{de-bert}  & 0.295 & 0.936 & 0.639 & 0.165 \\
 \textsc{me-bert}  & 0.323 & 0.968 & 0.687 & 0.109 \\
 \textsc{de-hybrid-e}  & 0.306 & 0.951 & 0.659 & 0.105  \\
 \textsc{me-hybrid-e}  &\textbf{0.336} & \textbf{0.977} & \textbf{0.706} & {0.051} \\
\midrule
 \multicolumn{5}{c}{Document Retrieval} \\  
 \midrule
 Base-Indri & 0.192 & 0.785 & 0.517 & \textbf{0.002} \\
 \textsc{de-bert}  & - & 0.841 & 0.510 & 0.188\\
 \textsc{me-bert}  & - & 0.877 & 0.588 & 0.109 \\
 \textsc{de-hybrid-e}  & 0.287 & 0.890 & 0.595 & 0.084  \\
 \textsc{me-hybrid-e}  & 0.310 & \textbf{0.914} & \textbf{0.610} & 0.063 \\
\bottomrule
\end{tabularx}
}
\caption{{\small  Test set first-pass retrieval results on the passage and document TREC 2019 DL evaluation as well as \msmarco eval MRR@10 (passage) and MRR@100 (document) under MRR(MS).}}
  \label{tab:maintest}
\end{table}

%% file: figures/MSMARCO_at_depth_figure.tex
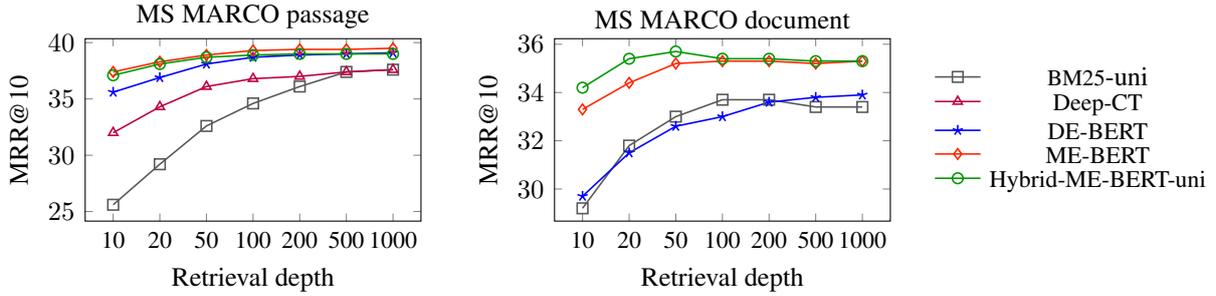
\begin{figure*}


\begin{tikzpicture}
    \pgfplotsset{footnotesize,samples=10}
    \begin{groupplot}[group style = {group size = 2 by 1, horizontal sep = 50pt}, width = 6.0cm, height = 4cm]
    \nextgroupplot[
            xticklabels from table={msmarco_sizes.dat}{Label},
            xticklabel style={align=center,font=\footnotesize},
            legend style = { column sep = 10pt, legend columns = -1, legend to name = grouplegend,},
            legend columns=2,
            every tick label/.append style={font=\small},
            legend style={column sep=0.2cm},
            legend style={anchor=south},
            legend style={at={(0.5, 0.0), anchor=south}, font=\large},
            ymin=25.5,ymax=39,
            enlarge y limits=true,
            title={{MS MARCO passage}},
            xlabel={Retrieval depth}, 
            ylabel={MRR@10},
            ]
             \addmsmarcobm{figures/data/results/MSMARCO_depth.csv};\label{addmsmarcobm}
            \addmsmarcodebert{figures/data/results/MSMARCO_depth.csv};\label{addmsmarcodebert}
             \addmsmarcomebert{figures/data/results/MSMARCO_depth.csv};\label{addmsmarcomebert}
             \addmsmarcohybrid{figures/data/results/MSMARCO_depth.csv};\label{addmsmarcohybrid}
             \addmsmarcodeepct{figures/data/results/MSMARCO_depth.csv};\label{addmsmarcodeepct}        

\coordinate (top) at (rel axis cs:0,1);
 \nextgroupplot[
            xticklabels from table={msmarco_sizes.dat}{Label},
            xticklabel style={align=center,font=\footnotesize},
            every tick label/.append style={font=\small},
            ymin=29.3,ymax=35.6,
            enlarge y limits=true,
            title={{MS MARCO document}},
            xlabel={Retrieval depth}, 
            ylabel={MRR@10},
            ]
             \addmsmarcobm{figures/data/results/MSMARCO_document_depth.csv};
            \addmsmarcodebert{figures/data/results/MSMARCO_document_depth.csv};
             \addmsmarcomebert{figures/data/results/MSMARCO_document_depth.csv};
             \addmsmarcohybrid{figures/data/results/MSMARCO_document_depth.csv};

\coordinate (bot) at (rel axis cs:1,0);

    \end{groupplot}
    
    \path (top)--(bot) coordinate[midway] (group center);
        
  \matrix[
      matrix of nodes,
      anchor=west,
      inner sep=0.14em,
    ]at([right=1em,inner sep=0pt]group center -| current bounding box.east)
    { \ref{addmsmarcobm}&{\small \bmone}  \\
            \ref{addmsmarcodeepct}&{\footnotesize{Deep-CT}}  \\
    \ref{addmsmarcodebert}&{\footnotesize{DE-BERT}}  \\
        \ref{addmsmarcomebert}&{\footnotesize{ME-BERT}}  \\
                \ref{addmsmarcohybrid}&\footnotesize{Hybrid-ME-BERT-uni}  \\
};

\end{tikzpicture}

                \caption{\small MRR@10 when reranking at different retrieval depth (10 to 1000 candidates) for \msmarco.}
                \label{fig:msmarco_depth}
\end{figure*}

%% file: figures/time_analysis.tex
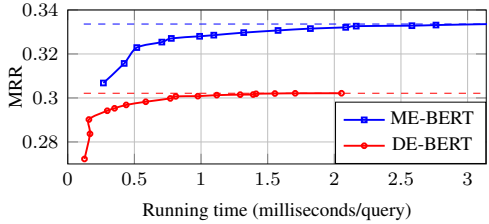
\begin{figure}[t]
\scalebox{.9}{
\begin{tikzpicture}
\begin{axis}[
    	width=1\columnwidth,
	    height=0.5\columnwidth,
	    legend style={at={(1, 0)},anchor=south east,font=\scriptsize},
	    mark options={mark size=1},
		font=\scriptsize,
		xlabel near ticks,
		ylabel near ticks,
	    xmin=0,xmax=3.14,
	    xtick={0,0.5,1,1.5,2,2.5,3},
   		ymin=0.27, ymax=0.34,
   	 	ymajorgrids=true, 
    	xmajorgrids=true, 
    	xlabel=Running time (milliseconds/query),
        ylabel=MRR,
    	ylabel style={yshift=-1.5ex,}]
 
\draw [dashed,blue] (0.12,0.3336) -- (3.14,0.3336);   
\draw [dashed,red] (0.12,0.30214) -- (3.14,0.30214);    

    \addplot[smooth,mark=square,blue,thick] plot coordinates {
(0.2674,0.3068)
(0.4233524355,0.3157)
(0.518008596,0.3229)
(0.7059139645,0.3254)
(0.7769924522,0.32707)
(0.989449195,0.32803)
(1.095788014,0.32855)
(1.320117217,0.3297)
(1.577872431,0.33069)
(1.820604542,0.33153)
(2.085491784,0.33209)
(2.164251312,0.33265)
(2.582090666,0.33285)
(2.762820728,0.33314)
(3.147905126,0.33357)
    };
    \addlegendentry{ME-BERT}
    
    \addplot[smooth,mark=o,red,thick] plot coordinates {
(0.1254527221,0.27227)
(0.1669169054,0.2837)
(0.1587106017,0.2902)
(0.2963696275,0.29418)
(0.3510988539,0.2953)
(0.4384527221,0.29683)
(0.5852292264,0.29827)
(0.7696031519,0.29979)
(0.8104942693,0.30072)
(0.9761575931,0.30082)
(1.117959885,0.301282)
(1.293839542,0.301503)
(1.391088825,0.301555)
(1.41272063,0.3019)
(1.554297994,0.30202)
(1.706548711,0.30212)
(2.0537,0.30214)
    };
    \addlegendentry{DE-BERT}
    
    \end{axis}
    
\end{tikzpicture}
}
\caption{\small{Quality/running time tradeoff for \textsc{de-bert} and \textsc{me-bert} on the \msmarco passage dev set. Dashed lines show quality with exact search.} 
\label{fig:timeanalysis}}
\end{figure}

%% file: related.tex
\section{Related work}

\newtext{We have mentioned research on improving the accuracy of retrieval models throughout the paper. Here we focus on work related to our central focus on the capacity of dense dual encoder representations relative to sparse bags-of-words.} 

In compressive sensing it is possible to recover a bag of words vector $x$ from the projection $Ax$ for suitable $A$. Bounds for the sufficient dimensionality of isotropic Gaussian projections~\cite{candes2005decoding,arora2018compressed} are more pessimistic than the bound described in \autoref{sec:theory}, but this is unsurprising because the task of recovering bags-of-words from a compressed measurement is strictly harder than recovering inner products.

\newcite{subramani2019can} ask whether it is possible to exactly recover sentences (token sequences) from pretrained decoders, using vector embeddings that are added as a bias to the decoder hidden state. Because their decoding model is more expressive (and thus more computationally intensive) than inner product retrieval, the theoretical \newtextfinal{issues examined} here do not apply. Nonetheless, \citeauthor{subramani2019can} empirically observe a similar dependence between sentence length and embedding size. 
\newcite{wieting2019no} represent sentences as bags of random projections, finding that high-dimensional projections ($\embsize = 4096$) perform nearly as well as trained encoding models. \commentout{ such as SkipThought~\citep{kiros2015skip} and InferSent~\cite{conneau2017supervised}.} 
These empirical results provide further empirical support for the hypothesis that bag-of-words vectors from real text are ``hard to embed'' in the sense of \newcite{larsen2017optimality}.
Our contribution is to systematically explore the relationship between document length and encoding dimension, focusing on the case of exact inner product-based retrieval.  We leave the combination of representation learning and approximate retrieval for future work. 

%% file: conclusion.tex
\section{Conclusion}
Transformers perform well on an unreasonable range of problems in natural language processing.
Yet the computational demands of large-scale retrieval push us to seek other architectures: cross-attention over contextualized embeddings is too slow, but dual encoding into fixed-length vectors may be insufficiently expressive, sometimes failing even to match the performance of sparse bag-of-words competitors. 
We have used both theoretical and empirical techniques to characterize the fidelity of fixed-length dual encoders, focusing on the role of document length. Based on these observations, we propose hybrid models that yield strong performance while maintaining scalability.

%% file: acknowledgments.tex
\paragraph{Acknowledgments}
We thank 
Ming-Wei Chang,
Jon Clark, William Cohen,
Kelvin Guu,
Sanjiv Kumar, Kenton Lee, Jimmy Lin, Ankur Parikh,
Ice Pasupat, Iulia Turc, 
William A. Woods,
Vincent Zhao,
and the anonymous reviewers
for helpful discussions of this work.

%% file: proofs.tex
\section{Proofs}
\subsection{Lemma \ref{lem:ranking-error-rp}}
\label{sec:proof_RP_error_bound}
\begin{proof} 
For both distributions of embeddings, the error on the squared norm can be bounded with high probability~\citep[][Lemma 5.1]{achlioptas2003database}:
\begin{equation}
\begin{split}
&    \Pr(\left| ||Ax||^2 - ||x||^2 \right| > \epsilon||x||^2 )\\
&    < 2 \exp (-\frac{\embsize}{2}(\epsilon^2/2 - \epsilon^3/3) ).
\end{split}
\end{equation}
This bound implies an analogous bound on the absolute error of the inner product~\cite[][corollary 19]{ben2002limitations},
\begin{eqnarray}
&& \hspace{-0.5cm} \Pr(|\iprod{Ax}{Ay} - \iprod{x}{y}| \geq
\frac{\epsilon}{2}(||x||^2 + ||y||^2)) \nonumber \\
& \leq & 4 \exp (-\frac{\embsize}{2}(\epsilon^2/2 - \epsilon^3/3)).
\label{eq:bd-bound}
\end{eqnarray}

Let $\overline{q} = q / ||q||$ and $\overline{d} = (d_1 - d_2) / ||d_1 - d_2||.$ Then $\nmarg(q, d_1, d_2) = \iprod{\overline{q}}{\overline{d}}$. A ranking error occurs if and only if $\iprod{A\overline{q}}{A\overline{d}} \leq 0$, which implies $|\iprod{A\overline{q}}{A\overline{d}} - \iprod{\overline{q}}{\overline{d}}| \geq \epsilon$. By construction $||\overline{q}|| = ||\overline{d}|| = 1$, so the probability of an inner product distortion $\geq \epsilon$ is bounded by the right-hand side of (\ref{eq:bd-bound}).
\end{proof}

\subsection{Corollary \ref{cor:quad-bound}}
\label{sec:proof-quad-bound}
\begin{proof}
We have $\epsilon = \nmarg(q, d_1, d_2) = \iprod{\overline{q}}{\overline{d}} \leq 1$ by the Cauchy-Schwarz inequality. For $\epsilon \leq 1$, we have $\epsilon^2/6 \leq \epsilon^2/2 - \epsilon^3/3$. We can then loosen the bound in (\ref{eq:bd-bound-in-lemma}) to $\beta \leq 4 \exp ( - \frac{\embsize}{2}\frac{\epsilon^2}{6})$. Taking the natural log yields $\ln \beta \leq \ln 4 - \epsilon^{2} \embsize / 12$, which can be rearranged into $\embsize \geq 12 \epsilon^{-2} \ln \frac{4}{\beta}.$
\end{proof}

\subsection{Lemma \ref{lem:recall-at-r}}
\label{sec:proof-recall-at-r}

\begin{proof}For convenience define $\nmarg(d_2) = \nmarg(q, d_1, d_2)$.
Define $\epsilon$ as in the theorem statement, and
$
{\cal D}_{\epsilon} = \{d_2 \in {\cal D}: \nmarg(q, d_1, d_2) \geq \epsilon\}.
$
We have
\begin{align*}
& \Pr(R \geq \recallthresh_0) \leq \Pr(\exists d_2 \in {\cal D}_{\epsilon}: A q_1 \leq A q_2)\\
&\leq \sum_{d_2 \in {\cal D}_{\epsilon}} 4 \exp ( - \frac{k}{2} (\nmarg(d_2)^2/2 - \nmarg(d_2)^3/3))\\
&\leq 4 |{\cal D}_{\epsilon}| \exp ( - \frac{k}{2} (\epsilon^2/2 - \epsilon^3/3)).
\end{align*}
The first inequality follows because the event $R \geq \recallthresh_0$ implies the event $\exists d_2 \in {\cal D}_{\epsilon}: A q_1 \leq A q_2$. 
The second inequality follows by a combination of Lemma 1 and the union bound. 
The final inequality follows because for any $d_2 \in {\cal D}_{\epsilon}$, $\mu(q, d_1, d_2) \geq \epsilon$. 
The theorem follows because $|{\cal D}_{\epsilon}| = |{\cal D}| - \recallthresh_0 + 1$.
\end{proof}

\subsection{Corollary~\ref{cor:boolean-inner-product}}
\label{sec:proof-boolean-ip}
\begin{proof}
For the retrieval function $\max_{d} \iprod{q}{d}$, the minimum non-zero unnormalized margin ${\iprod{q}{d_1} - \iprod{q}{d_2}}$
is $1$ when $q$ and $d$ are Boolean vectors. Therefore the normalized margin has lower bound $\nmarg(q, d_1, d_2) \geq 1 / (||q|| \times ||d_1 - d_2||)$. For non-negative $d_1$ and $d_2$ we have $||d_1 - d_2|| \leq \sqrt{||d_1||^2 + ||d_2||^2} \leq \sqrt{2 L_D}.$
Preserving a normalized margin of $\epsilon = (2 L_Q L_D)^{-\frac{1}{2}}$ is therefore sufficient to avoid any pairwise errors. By plugging this value into Corollary~\ref{cor:quad-bound}, we see that setting $\embsize \geq 24 L_Q L_D \ln \frac{4}{\beta}$ ensures that the probability of any pairwise error is $\leq \beta$.
\end{proof}

\subsection{Theorem \ref{thm:multivector}}
\label{sec:proof-multivector}
\begin{proof}
Recall that $\nmarg(q, d_1, d_2) = \frac{\iprod{q}{d_1 - d_2}}{||q|| \times ||d_1 - d_2||}$. By assumption we have $\iprod{q}{d_1^{(i)}} = \iprod{q}{d_1} $ and $\max_j \iprod{q}{d_2^{(j)}} \leq \iprod{q}{d_2}$, implying that
\begin{equation}
    \label{eq:thm1-numerator}
    \iprod{q}{d_1^{(i)} - d_2^{(i)}} \geq \iprod{q}{d_1 - d_2}
    \end{equation}
In the denominator, we expand  $||d_1 - d_2|| = ||(d_1^{(i)} - d_2^{(i)}) + (d_1^{(\neg i)} - d_2^{(\neg i)})||$, where $d^{(\neg i)} = \sum_{j \neq i} d^{(j)}.$ Plugging this into the squared norm,
\begin{align}
\notag
&    ||d_1 - d_2||^2\\
& = ||(d_1^{(i)} - d_2^{(i)}) + (d_1^{(\neg i)} - d_2^{(\neg i)})||^2\\
    \begin{split}
    & = 
        ||d_1^{(i)} - d_2^{(i)}||^2
    + ||d_1^{(\neg i)} - d_2^{(\neg i)}||^2\\
    &{} \quad + 2\iprod{d_1^{(i)} - d_2^{(i)}}{d_1^{(\neg i)} - d_2^{(\neg i)}}
    \end{split}\\
    \label{eq:thm1-proof-inner-product-is-zero}
    & = ||d_1^{(i)} - d_2^{(i)}||^2
    + ||d_1^{(\neg i)} - d_2^{(\neg i)}||^2\\
    & \geq ||d_1^{(i)} - d_2^{(i)}||^2.
\label{eq:thm1-denominator}
\end{align}
The inner product $\iprod{d_1^{(i)} - d_2^{(i)}}{d_1^{(\neg i)} - d_2^{(\neg i)}} = 0$ because the segments are orthogonal. The combination of (\ref{eq:thm1-numerator}) and (\ref{eq:thm1-denominator}) completes the theorem.
\end{proof}

%% file: tables/ICT_by_length.tex
 \begin{table}[!t]
   \centering
   \small
   \scalebox{.75}{
   \begin{tabular}{l @{\hspace{0.1cm}} | c  c  c  c | c c c c}
     \toprule
     \multicolumn{1}{c}{Model} & \multicolumn{4}{c}{Reranking} & \multicolumn{4}{c}{Retrieval}\\
     \cmidrule{1-9}
     Passage length & {50} & {100} & {200} & {400} & {50} & {100} & {200} & {400}\\
    \midrule
      \multicolumn{9}{c}{ICT task (MRR@10)} \\
      \midrule
   \small{\crossatt} & 99.9 & 99.9 & 99.8 & 99.6 & - & - & - & -\\ 
\footnotesize{\hybrid}-{\qonedeight}-\footnotesize{uni} & - & - & - & - & 98.2 & 97.0 & 94.4 & 91.9  \\ 
  \footnotesize{\hybrid}-{\qonedeight}-\footnotesize{bi} & - & - & - & - & 99.3 & 99.0 & 97.3 & 96.1 \\
     \small{\qonedeight}-\footnotesize{768} & 98.0 & 96.7 & 92.4 & 89.8 & 96.8 & 96.1 & 91.1 & 85.2 \\ 
     \small{\qonedeight}-\footnotesize{64} & 96.3 & 94.2 & 89.0 & 83.7 & 92.9 & 91.7 & 84.6 & 72.8 \\   
     \small{\de}-\footnotesize{768} & 91.7 & 87.8 & 79.7 & 74.1 & 90.2 & 85.6 & 72.9 & 63.0 \\
     \small{\de}-\footnotesize{512} & 91.4 & 87.2 & 78.9 & 73.1 & 89.4 & 81.5 & 66.8 & 55.8 \\
     \small{\de}-\footnotesize{128} & 90.5 & 85.0 & 75.0 & 68.1 & 85.7 & 75.4 & 58.0 & 47.3 \\ 
    \small{\de}-\footnotesize{64} & 88.8 & 82.0 & 70.7 & 63.8 & 82.8 & 68.9 & 48.5 & 38.3 \\ 
    \small{\de}-\footnotesize{32} & 83.6 & 74.9 & 62.6 & 55.9 & 70.1 & 53.2 & 34.0 & 27.6\\
    \bmone & 92.1 & 88.6 & 84.6 & 81.8 & 92.1 & 88.6 & 84.6 & 81.8 \\
   \bmtwo & 98.0 & 97.1 & 95.9 & 94.5 & 98.0 & 97.1 & 95.9 & 94.5\\ 
   \midrule
      \multicolumn{9}{c}{NQ (Recall@400 tokens)} \\
   \midrule
   \small{\crossatt} & 48.9 & 55.5 & 54.2 & 47.6 & - & - & - & - \\
\footnotesize{\hybrid}-{\qonedeight}-\footnotesize{uni} & - & - & - & - & 45.7 & 49.5 & 48.5 & 42.9 \\
     \small{\qonedeight}-\footnotesize{768} & 43.6 & 49.6 & 46.5 & 38.7 & 42.0 & 43.3 & 40.4 & 34.4 \\
     \small{\qonedeight}-\footnotesize{64} & 44.4 & 48.7 & 44.5 & 38.2 & 42.2 & 43.4 & 38.9 & 33.0\\ 
     \small{\de}-\footnotesize{768} & 42.9 & 47.7 & 44.4 & 36.6 & 44.2 & 44.0 & 40.1 & 32.2\\
     \small{\de}-\footnotesize{512} & 43.8 & 48.5 & 44.1 & 36.5 & 43.3 & 43.2 & 38.8 & 32.7\\
     \small{\de}-\footnotesize{128} & 42.8 & 45.7 & 41.2 & 35.7 & 38.0 & 36.7 & 32.8 & 27.0 \\ 
    \small{\de}-\footnotesize{64} & 42.6 & 45.7 & 42.5 & 35.4 & 37.4 & 35.1 & 32.6 & 26.6 \\ 
    \small{\de}-\footnotesize{32} & 42.4 & 45.8 & 42.1 & 34.0 & 36.3 & 34.7 & 31.0 & 24.9 \\ 
    \bmone & 30.1 & 35.7 & 34.1 & 30.1 & 30.1 & 35.7 & 34.1 & 30.1 \\
     \bottomrule
   \end{tabular}
   }
   \caption{{\small{Results on ICT task and NQ task (correspond to Fig. \ref{fig:unmasked} and Fig. \ref{fig:nq}). }}}
   \label{tab:ICT_NQ}
\end{table}